\documentclass[10pt,twocolumn,letterpaper]{article}

\usepackage{cvpr}
\usepackage{times}
\usepackage{epsfig}
\usepackage{graphicx}
\usepackage{amsmath}
\usepackage{amssymb}
\usepackage{multirow}
\usepackage[toc,page]{appendix}

\usepackage[pagebackref=true,breaklinks=true,letterpaper=true,colorlinks,bookmarks=false]{hyperref}
\cvprfinalcopy 

\def\cvprPaperID{} 

\ifcvprfinal\fi
\begin{document}

\title{Detailed 2D-3D Joint Representation for Human-Object Interaction}
\author{Yong-Lu Li,~Xinpeng Liu,~Han Lu,~Shiyi Wang,~Junqi Liu,~Jiefeng Li,~Cewu Lu\thanks{Cewu Lu is the corresponding author, member of Qing Yuan Research Institute and MoE Key Lab of Artificial Intelligence, AI Institute, Shanghai Jiao Tong University, China.}\\
Shanghai Jiao Tong University\\
{\tt\small yonglu\_li@sjtu.edu.cn,}
{\tt\small xinpengliu0907@gmail.com,}\\
{\tt\small \{sjtu\_luhan, shiywang, ljq435, ljf\_likit, lucewu\}@sjtu.edu.cn}
}

\maketitle
\thispagestyle{empty}

\begin{abstract}
   Human-Object Interaction (HOI) detection lies at the core of action understanding. Besides 2D information such as human/object appearance and locations, 3D pose is also usually utilized in HOI learning since its view-independence. However, rough 3D body joints just carry sparse body information and are not sufficient to understand complex interactions. Thus, we need detailed 3D body shape to go further. Meanwhile, the interacted object in 3D is also not fully studied in HOI learning. In light of these, we propose a detailed 2D-3D joint representation learning method. First, we utilize the single-view human body capture method to obtain detailed 3D body, face and hand shapes. Next, we estimate the 3D object location and size with reference to the 2D human-object spatial configuration and object category priors. Finally, a joint learning framework and cross-modal consistency tasks are proposed to learn the joint HOI representation. To better evaluate the 2D ambiguity processing capacity of models, we propose a new benchmark named Ambiguous-HOI consisting of hard ambiguous images. Extensive experiments in large-scale HOI benchmark and Ambiguous-HOI show impressive effectiveness of our method. Code and data are available at \url{https://github.com/DirtyHarryLYL/DJ-RN}.
\vspace{-0.5cm}
\end{abstract}

\section{Introduction}
Human-Object Interaction (HOI) detection  recently receives lots of attentions.
It aims at locating the active human-object and inferring the action simultaneously. 
As a sub-task of visual relationship~\cite{Lu2016Visual}, it can facilitate activity understanding~\cite{activitynet,pang2019deep,pang2020deep,shao2018find}, imitation learning~\cite{immitation}, \etc. 
\begin{figure}[!ht]
	\begin{center}
		\includegraphics[width=0.45\textwidth]{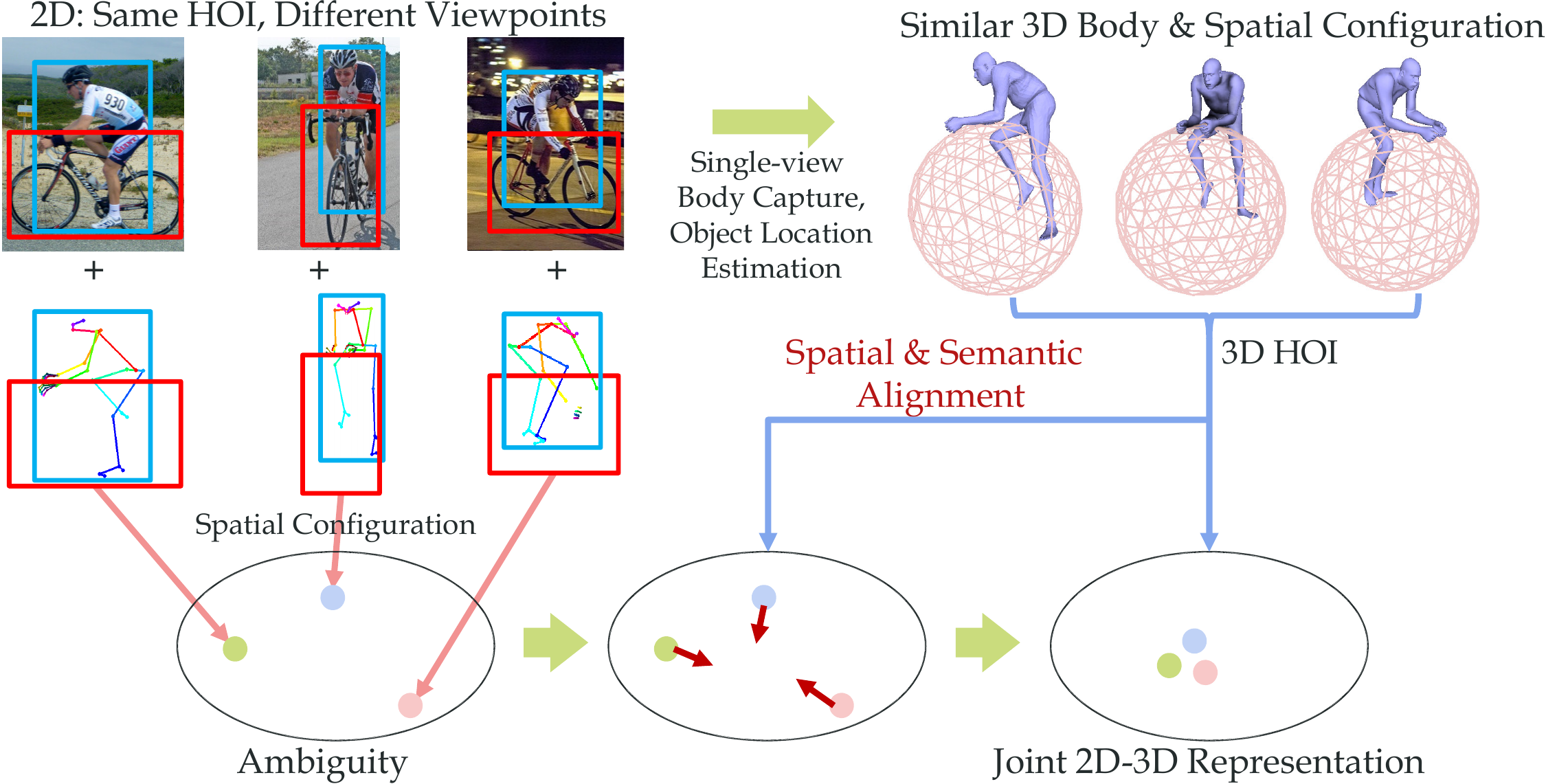}
	\end{center}
	\caption{HOI detection based on 2D may have ambiguities under various viewpoints. HOI representation in 3D is more robust. Thus, we estimate the 3D detailed human body and interacted object location and size to represent the HOI in 3D. Then we learn a joint 2D-3D representation to combine multi-modal advantages.}
	\label{Figure:insight}
\vspace{-0.5cm}
\end{figure}

What do we need to understand HOI? The possible answers are human/object appearance, spatial configuration, context, pose, \etc.
Among them, human body information often plays an important role, such as 2D pose~\cite{interactiveness,NoFrills,shangke,du2015hierarchical} and 3D pose~\cite{yao2013discovering,luvizon20182d}.
Because of the various viewpoints, 2D human pose~\cite{openpose} or segmentation~\cite{maskrcnn,xu2018srda,fang2019instaboost} often has ambiguities, \eg same actions may have very different 2D appearances and poses.
Although 3D pose is more robust, rough 3D body joints are not enough to encode essential geometric and meaningful patterns.
For example, we may need detailed hand shape to infer the action ``use a knife to cut'', or facial shape for ``eat and talk''. And body shape would also largely affect human posture.
In light of this, we argue that \textit{detailed 3D body} can facilitate the HOI learning.
Meanwhile, the object in HOI is also important, \eg ``hold an apple'' and ``hold the horse'' have entirely different patterns.
However, few studies considered how to embed 3D interacted objects in HOI.
The reasons are two-fold: first, it is hard to reconstruct objects because of the 6D pose estimation and diverse object shapes (detailed point cloud or mesh~\cite{chang2015shapenet,you2018prin}). 
Second, estimating 3D human-object spatial relationship is also difficult for single-view.

In this work, we propose a method to not only borrow essential discriminated clues from the detailed 3D body but also consider the 3D human-object spatial configuration.
First, we represent the HOI in 3D.
For human, we utilize the single-view human body capture~\cite{smplify-x} to obtain detailed human shape.
For object, referring to the 2D human-object spatial configuration and object category prior, we estimate its rough \textit{location} and \textit{size} through perspective projection, and use a hollow sphere to represent it.
Then, we put the 3D detailed human and object sphere into a normalized volume as the \textbf{3D HOI spatial configuration volume}, as shown in Fig.~\ref{Figure:insight}.
Next, we propose Detailed Joint Representation Network (DJ-RN), which consists of two feature extractors: a 2D Representation Network (2D-RN) and a 3D Representation Network (3D-RN).
Finally, we adopt several consistency tasks to learn the 2D-3D joint HOI representation.
In detail, we align the 2D spatial features according to more robust 3D spatial features. 
And we perform semantic alignment to ensure the cross-modal semantic consistency. 
To better embed the body posture, we estimate body part attentions in a 2D-3D joint way with consistency. That is if 2D features tell us the hands and head are important for ``work on laptop'', so will the 3D features.
DJ-RN is the \textit{first} joint learning method to utilize single-view 3D recover for {\bf HOI}.
It is a novel {\bf paradigm} instead of an ad-hoc model, and flexible to replace 2D/3D modules/extracted features.
We believe it would promote not only HOI learning but also action related tasks, \eg, image caption, visual reasoning.

To better evaluate the ability of processing 2D ambiguities, we propose a new benchmark named \textbf{Ambiguous-HOI}, which includes ambiguous examples selected from existing datasets like HICO-DET~\cite{hicodet}, V-COCO~\cite{vcoco}, OpenImage~\cite{openimages}, HCVRD~\cite{hcvrd}.
We conduct extensive experiments on widely-used HOI detection benchmark and Ambiguous-HOI. 
Our approach achieves significant improvements with 2D-3D joint learning.
The main contributions are as follows: 
1) We propose a 2D-3D joint representation learning paradigm to facilitate HOI detection. 
2) A new benchmark Ambiguous-HOI is proposed to evaluate the disambiguation ability of models.
3) We achieve state-of-the-art results on HICO-DET~\cite{hicodet} and Ambiguous-HOI.

\section{Related Work}
\noindent{\bf Human-Object Interaction Detection.} Recently, great progress has been made in HOI detection. 
Large-scale datasets~\cite{hicodet,vcoco,openimages} have been released to promote this field.
Meanwhile, lots of deep learning based methods~\cite{Gkioxari2017Detecting,gao2018ican,interactiveness,qi2018learning,NoFrills,shangke,peyre2018detecting} have been proposed. 
Chao~\etal~\cite{hicodet} proposed a multi-stream framework, which is proven effective and followed by subsequent works~\cite{gao2018ican,interactiveness}.
Differently, GPNN~\cite{qi2018learning} proposed a graph model and used message passing to address both image and video HOI detection.
Gkioxari~\etal~\cite{Gkioxari2017Detecting} adopted an action density map to estimate the 2D location of interacted objects.
iCAN~\cite{gao2018ican} utilized self-attention to correlate the human-object and context.
TIN~\cite{interactiveness} proposed an explicit interactiveness learning network to identify the non-interactive human-object pairs and suppress them in inference.
HAKE~\cite{li2019hake} proposes a novel hierarchical paradigm based on human body part states~\cite{lu2018beyond}.
Previous methods mainly relied on the visual appearance and human-object spatial relative locations, some of them~\cite{interactiveness} also utilized the 2D estimated pose. But the 2D ambiguity in HOI is not well studied before.

\noindent{\bf 3D Pose-based Action Recognition.} 
Recent deep learning based 3D pose estimation methods~\cite{hmr,fang2018learning,coarse} have achieved substantial progresses.
Besides 2D pose based action understanding~\cite{interactiveness,lee2017ensemble,liu2018recognizing,choutas2018potion,girdhar2017attentional,wang2013approach}, many works also utilized the 3D human pose~\cite{shahroudy2016ntu,du2015hierarchical,Yao2012Action,yao2013discovering,luvizon20182d,pham2019unified,ke2017new,liu2016spatio,activity2,activity3,activity4,activity5,huang2017deep}.
Yao~\etal~\cite{Yao2012Action} constructed a 2.5D graph with 2D appearance and 3D human pose, and selected exemplar graphs of different actions for the exemplar-based action classification.
In \cite{yao2013discovering}, 2D pose is mapped to 3D pose and the actions are classified by comparing the 3D pose similarity.
Luvizon~\etal~\cite{luvizon20182d} estimated the 2D/3D pose and recognized actions in a unified model from both image and video.
Wang~\etal~\cite{wang2013learning} used the RGB-D data to obtain the 3D human joints and adopted an actionlet ensemble method for HOI learning.
Recently, Pham~\etal~\cite{pham2019unified} proposed a multi-task model to operate 3D pose estimation and action recognition simultaneously from RGB video. 
Most 3D pose based methods~\cite{activity2,activity3,activity4,activity5,liu2016spatio,pham2019unified,wang2013learning,du2015hierarchical,ke2017new,huang2017deep} are using Recurrent Nerual Network (RNN) based framework for spatio-temporal action recognition, but few studies focus on the complex HOI understanding from single RGB image.

\noindent{\bf Single-view 3D Body Recover.} Recently the single-view human body capture and reconstruction methods~\cite{smplify-x,hmr,omran2018neural,pavlakos2018learning,bogo2016keep} have made great progresses. With the help of deep learning and large-scale scanned 3D human database~\cite{cmu,human3.6m,poseprior}, they are able to directly recover 3D body shape and pose from single RGB images.
SMPLify-X~\cite{smplify-x} is a holistic and efficient model that takes the 2D human body, face and hand poses as inputs to capture 3D body, face and hands. To obtain more accurate and realistic body shape, SMPLify-X~\cite{smplify-x} utilizes the Variational Human Body Pose prior (VPoser) trained on large-scale MoCap datasets, which carries lots of human body pose prior and knowledge.
It supports us to recover 3D detailed human body from HOI images and embed more body posture knowledge.

\section{Representing HOI in 3D}
Our goal is to learn the 2D-3D joint HOI representation, thus we need to first represent HOI in 3D.
Given a still image, we use object detection~\cite{faster-rcnn} and pose estimation~\cite{openpose} to obtain 2D instance boxes and human pose.
Next, we adopt the 3D human body capture~\cite{smplify-x} to estimate the 3D human body with above 2D detection (Sec.~\ref{sec:3d}), and estimate the object location and size in 3D to construct the 3D spatial configuration volume (Sec.~\ref{sec:volume}).

\subsection{Single-view 3D Body Capture}
\label{sec:3d}
\begin{figure}[!ht]
	\begin{center}
		\includegraphics[width=0.45\textwidth]{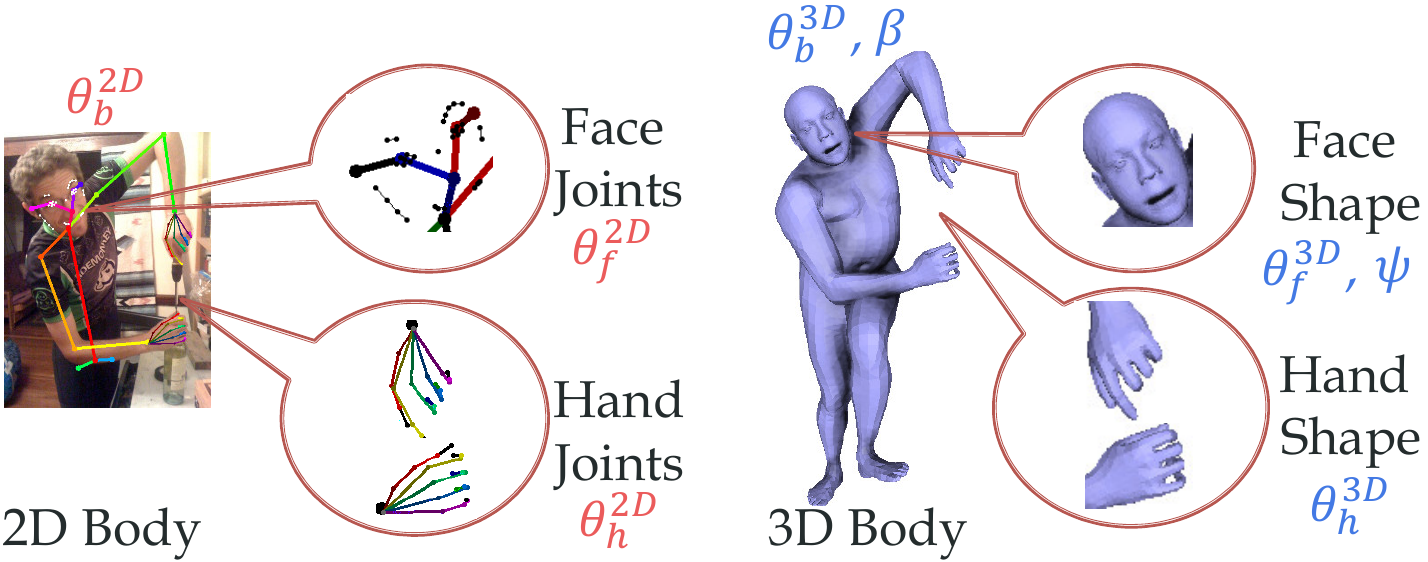}
	\end{center}
	\caption{We adopt detailed human information. We use OpenPose~\cite{openpose} and SMPLify-X~\cite{smplify-x} to estimate the 2D, 3D poses and shapes of face and hands. These information would largely help the HOI learning, especially on actions related to face and hands.}
	\label{Figure:detailed}
\vspace{-0.5cm}
\end{figure}
Rough 3D pose is not sufficient to discriminate various actions, especially the complex interactions with daily objects. 
Thus we need holistic and fine-grained 3D body information as a clue. 
To this end, we adopt a holistic 3D body capture method~\cite{smplify-x} to recover detailed 3D body from single RGB images.
Given the 2D detection of image $I$, \ie, 2D human and object boxes $b_h$ and $b_o$, 2D human pose $\theta^{2D}=\{\theta^{2D}_b, \theta^{2D}_f, \theta^{2D}_h\}$ (main body joints $\theta^{2D}_b$, jaw joints $\theta^{2D}_f$ and finger joints $\theta^{2D}_h$ in Fig.~\ref{Figure:detailed}).
We input them into SMPLify-X~\cite{smplify-x} to recover 3D human estimations, \ie, fitting the SMPL-X~\cite{smplify-x} model to $I$ and $\theta^{2D}$.
Then we can obtain the optimized shape parameters $\{\theta^{3D},\beta,\psi\}$ by minimizing the body pose, shape objective function, where $\theta^{3D}$ are pose parameters and $\theta^{3D}=\{\theta^{3D}_b, \theta^{3D}_f, \theta^{3D}_h\}$, $\beta$ are body, face and hands shape parameters, $\psi$ are facial expression parameters. 
The template body mesh is finally blended and deformed to fit the target body posture and shape in images.
With function $M(\theta^{3D},\beta,\psi): \mathbb{R}^{|\theta^{3D}|\times|\beta|\times|\psi|} \to \mathbb{R}^{3N}$, we can directly generate the 3D body mesh according to the estimated $\{\theta^{3D},\beta,\psi\}$ from images and utilize it in the next stage, some examples are shown in Fig.~\ref{Figure:detailed}. 

\subsection{3D Spatial Configuration Volume}
\label{sec:volume}
After obtaining the 3D body, we further represent HOI in 3D, \ie estimate the 3D object location and size.
For robustness and efficiency, we do not reconstruct the object shape, but use a \textit{hollow sphere} to represent it. Thus we can avoid the difficult 6D pose estimation under the circumstance of single-view and various categories.
Our procedure has \textit{two stages}: 
1) locating the sphere center on a plane according to the camera perspective projection, 2) using the prior object size and human-object distance to estimate the depth of the sphere. 
For each image, we adopt the estimated camera parameters from SMPLify-X~\cite{smplify-x}, where focal length $f$ is set to a fixed value of 5000, and the camera distortions are not considered.
For clarification, the camera optical center is expressed as $C(t_1, t_2, t_3)$ in the world coordinate system, and the object sphere center is $O(x_O, y_O, z_O)$.
\begin{figure}[!ht]
	\begin{center}
		\includegraphics[width=0.45\textwidth]{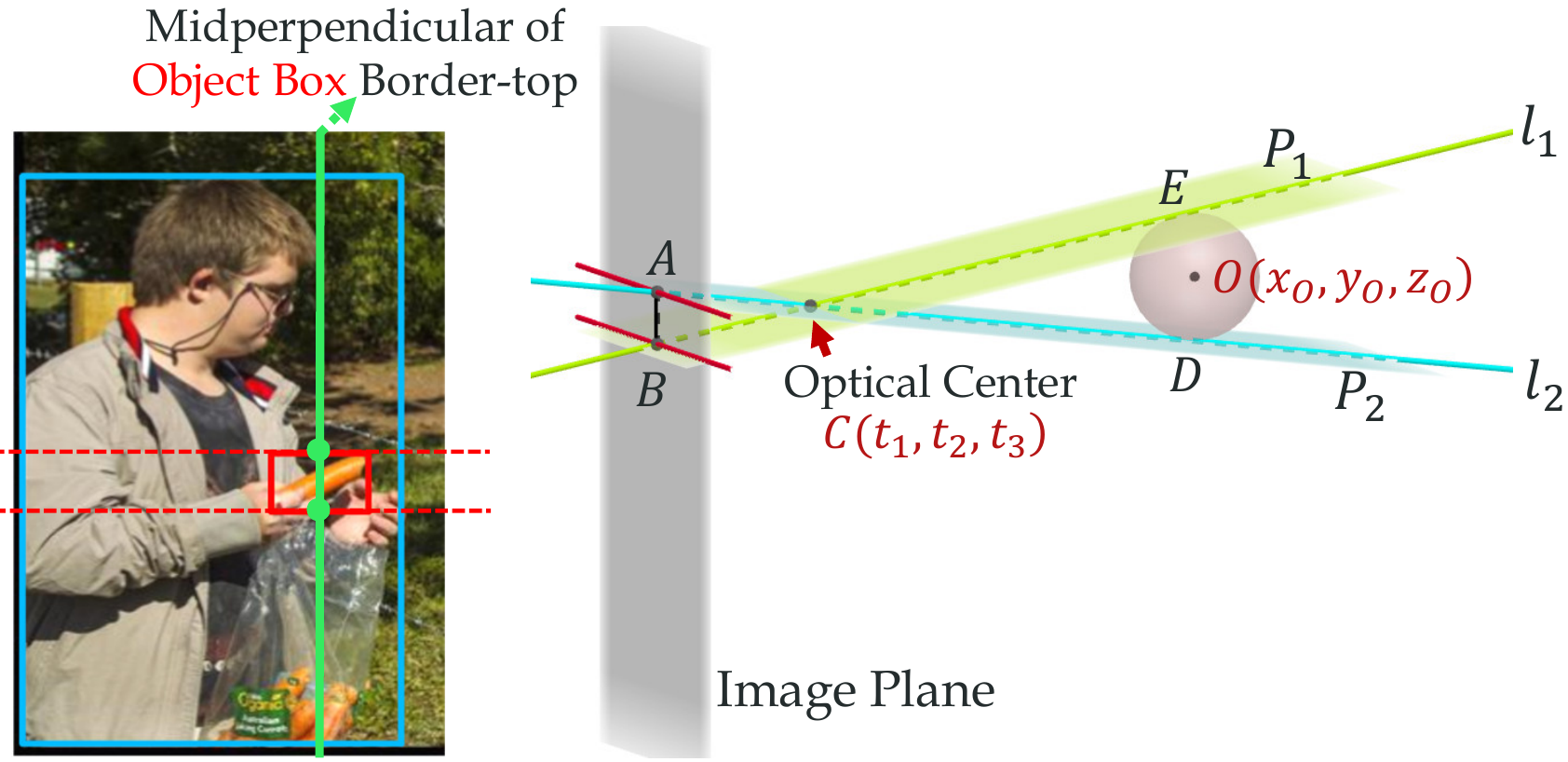}
	\end{center}
	\caption{Object location estimation. 
	Given prior radius $r$, we can get the sphere center location by solving projection equations, which restricts the sphere to be tangent to plane $P_1$ and $P_2$, and assures the sphere center falls on plane $P_{ABC}$.}
	\label{Figure:volume_consturct}
\vspace{-0.5cm}
\end{figure}

\noindent {\bf Object Sphere Center.}
As shown in Fig.~\ref{Figure:volume_consturct}, we assume that $O$ is projected to the midperpendicular of object box border-top, indicating the sphere center falls on plane $P_{ABC}$.  
And we suppose the highest and lowest visible points of the sphere are projected to the object box border-bottom and border-top respectively. 
Then we can get two tangent planes of the sphere: $P_1$ (contains points B, C, E) and $P_2$ (contains points A, C, D) as shown in Fig.~\ref{Figure:volume_consturct}.
$P_1$ and $P_2$ intersect with $P_{ABC}$, restricting a region on $P_{ABC}$ where the sphere center may locate. To get the depth of sphere center, we need to know the sphere radius, \ie, $r=|\overrightarrow{OD}|=|\overrightarrow{OE}|$.

\noindent {\bf Object Sphere Radius.}
As for the sphere radius, we determine it by both considering the object box relative size (to the 2D human box) and the object category prior. 
With object detection in the first step, we can obtain the object category $j$. Thus, we can set a rough object size according to Wikipedia and daily life experience. In practice, we set prior sizes for COCO 80 objects~\cite{coco} to suit the HICO-DET setting~\cite{hicodet}. 
First, for small objects or objects with similar size along different axes (\eg ball, table), we define the \textit{prior object scale ratio} between the sphere radius and the human shoulder width.
Second, for objects that are usually partly seen or whose projection is seriously affected by the 6D pose (\eg boat, skis), we use the relative scale ratio of the human and object boxes as the referenced ratio. 
The estimated sphere center is denoted as $\hat{O}(\hat{x}_O, \hat{y}_O, \hat{z}_O)$.

The sphere depth is very sensitive to the radius and may make the sphere away from human. 
Thus, we regularize the estimated depth $\hat{z}_c$ using the maximum and minimum depth $z_H^{max}, z_H^{min}$ of the recovered human. 
We define prior object \textit{depth regularization factor} $\Gamma=\{[\gamma^{min}_i, \gamma^{max}_i]\}^{80}_{i=1}$ for COCO objects~\cite{coco}. 
Specifically, with pre-defined depth bins (very close, close, medium, far, very far), we invite fifty volunteers from different backgrounds to watch HOI images and choose the degree of the object relative depth to the human.
We then use their votes to set the empirical regularization factors $\Gamma$. 
For estimated $\hat{O}(\hat{x}_O, \hat{y}_O, \hat{z}_O)$, if $\hat{z}_O$ falls out of $[\gamma_j^{min}z_H^{min}, \gamma_j^{max}z_H^{max}]$, we shift $\hat{O}$ to $(\hat{x}_O, \hat{y}_O, \gamma_j^{max}z_H^{max})$ or $(\hat{x}_O, \hat{y}_O, \gamma_j^{min}z_H^{min})$, depending on which is closer to $\hat{O}$.
Size and depth priors can effectively restrict the error boundaries.
Without them, 3D volume would have large deviation and degrade performance.

\noindent {\bf Volume Formalization.}
Next, we perform translations to align different configurations in 3D. 
First, we set the coordinate origin as the human pelvis. The direction of gravity estimated is kept same with the negative direction of the z-axis, and the line between two human shoulder joints is rotated to be parallel to the x-axis.
Then, we down-sample the 3D body to 916 points and randomly sample 312 points on \textit{spherical surface}. The hollow sphere can keep the body information of the interacted body parts within the sphere.
We then normalize the whole volume by setting unit length as the distance between the pupil joints.
At last, we can obtain a normalized 3D volume including 3D body and object sphere, which not only carries essential 3D action information but also 3D human-object spatial configuration.

\section{2D-3D Joint Learning}
\label{sec:joint-learning}
In this section, we aim to learn the joint representation. 
To this end, we propose Detailed Joint Representation Network (DJ-RN), as seen in Fig.~\ref{Figure:overview}. 
DJ-RN has two modules: 2D Representation Network (2D-RN) and 3D Representation Network (3D-RN). 
We use them to extract features from two modalities respectively (Sec.~\ref{sec:extraction}, \ref{sec:3d-extract}).
Then we align 2D spatial feature with 3D spatial feature (Sec.~\ref{sec:sp-align}), and use body part attention consistency (Sec.~\ref{sec:att-align}) and semantic consistency (Sec.~\ref{sec:semantic-align}) to guide the learning .

\subsection{2D Feature Extraction}
\label{sec:extraction}
\begin{figure*}[!ht]
	\begin{center}
		\includegraphics[width=0.9\textwidth]{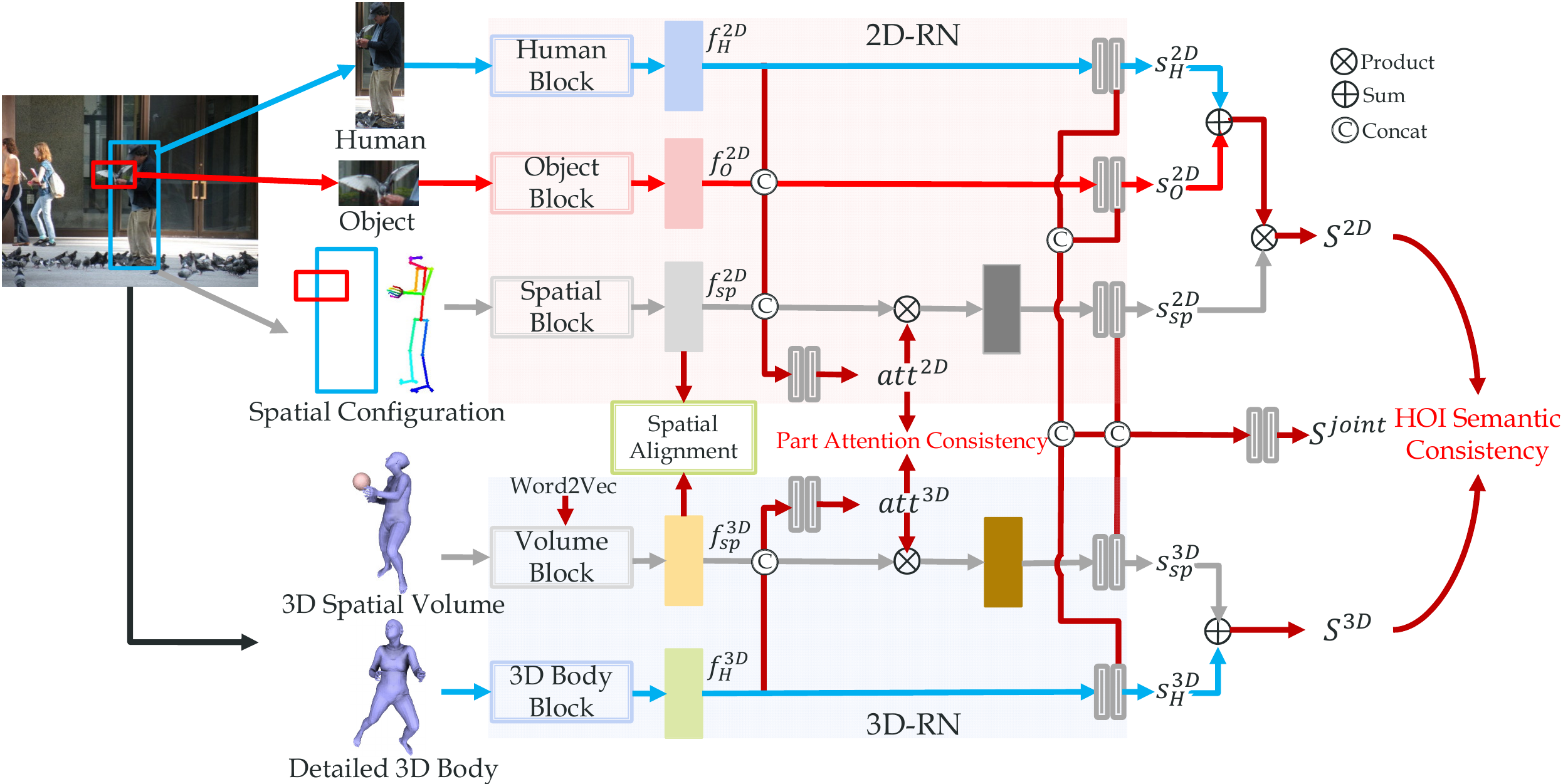}
	\end{center}
	\caption{Overview of DJ-RN. The framework consists of two main modules, named 2D Representation Network (2D-RN) and 3D Representation Network (3D-RN). They extract HOI representations from 2D and 3D information respectively. Hence, we can use spatial alignment, part attention consistency and semantic consistency to learn a joint 2D-3D representation for HOI learning.}
	\label{Figure:overview}
	\vspace{-0.5cm}
\end{figure*}

2D-RN is composed of human, object, and spatial streams following~\cite{hicodet,gao2018ican,interactiveness}. 
Within each stream, we adopt different blocks to take in 2D information with different properties and extract corresponding features (Fig.~\ref{Figure:overview}).

\noindent{\bf Human/Object Block.} 
Human and object streams mainly utilize visual appearance. We use a COCO~\cite{coco} pre-trained Faster-RCNN~\cite{faster-rcnn} to extract ROI pooling features from detected boxes. 
To enhance the representation ability, we adopt the iCAN block~\cite{gao2018ican} which computes the self-attention via correlating the context and instances, and obtain the human feature $f^{2D}_H$ and object feature $f^{2D}_O$.

\begin{figure}[!ht]
	\begin{center}
		\includegraphics[width=0.45\textwidth]{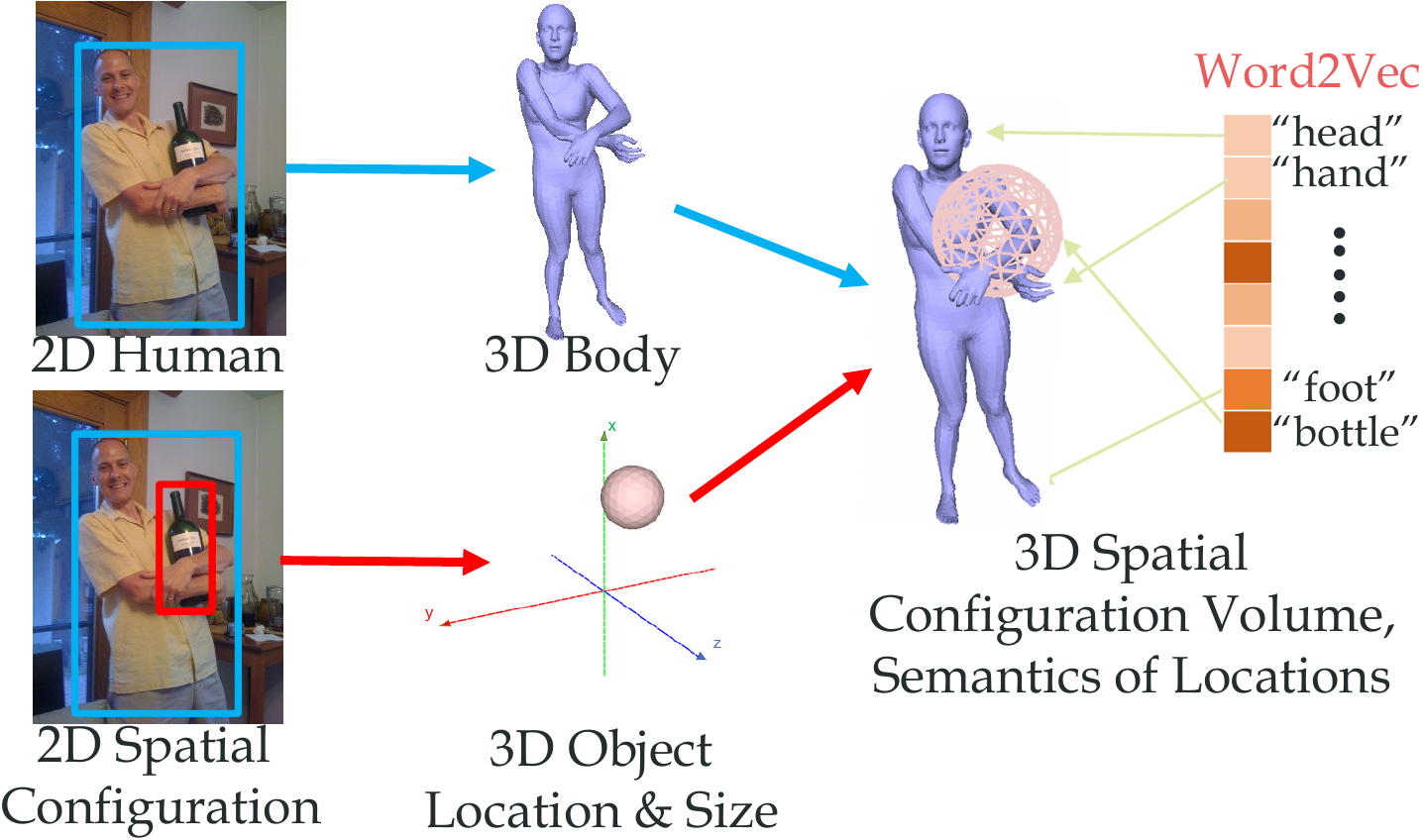}
	\end{center}
	\caption{3D spatial configuration volume. After 3D body capture, we use 2D boxes, estimated camera parameters and object category prior to estimate the 3D object location and size, and then put 3D human and object together in a normalized volume. We also pair the 3D location with semantic knowledge (Sec.~\ref{sec:3d-extract}).}
	\label{Figure:volume}
\vspace{-0.3cm}
\end{figure}

\noindent{\bf Spatial Block.}
Although appearance carries important clues, it also imports noise and misleading patterns from various viewpoints. Thus human-object spatial configuration can be used additionally to provide discriminative features~\cite{hicodet,gao2018ican,interactiveness}.
Spatial stream mainly considers the 2D human-object relative locations. 
We input the 2D pose map and spatial map~\cite{interactiveness} to the spatial block, which consists of convolution and fully-connected (FC) layers to extract the spatial feature $f^{2D}_{sp}$.
The spatial map consists of two channels, human and object maps, which are all $64 \times 64$ and generated from the human and object boxes. The value is 1 in the box and 0 elsewhere.
The pose map consists of 17 joint heatmaps of size $64 \times 64$ from OpenPose~\cite{openpose}.

\subsection{3D Feature Extraction}
\label{sec:3d-extract}
3D-RN contains a 3D spatial stream with volume block which takes in the 3D spatial configuration volume, and a 3D human stream with 3D body block to encode 3D body.

\noindent{\bf Volume Block.}
In 3D spatial stream, we adopt PointNet~\cite{pointnet} to extract 3D spatial feature $f^{3D}_{sp}$. 
We first pre-train it to segment the human and object points in the generated 3D spatial configuration volume.
Thus it can learn to discriminate the geometric difference and shape of human and object. 
Then we use it to extract features from 3D spatial volume point cloud. 
To further embed the \textbf{semantic} information of 3D \textbf{locations}, we \textit{pair} the spatial feature with the corresponding semantics, \ie, the word embedding of object or body part category. 
We first divide the volume point cloud into 18 sets: 17 part sets and an object sphere set. 
Then, for the feature of part set, we concatenate it with PCA reduced word embedding~\cite{word2vec} of part name (\eg ``hand'').
Similarly, for the feature of the sphere set, we concatenate it with the object category word embedding (\eg ``bottle''), as seen in Fig.~\ref{Figure:volume}.
The concatenated feature is used as $f^{3D}_{sp}$.

\noindent{\bf 3D Body Block.}
In 3D body block, we extract features based on SMPL-X~\cite{smplify-x} parameters: joint body, face and hands shape $\beta$, face expression $\psi$ and pose $\theta^{3D}$, consisting of jaw joints $\theta^{3D}_f$, finger joints $\theta^{3D}_h$ and body joints $\theta^{3D}_b$. 
For body shape and expression, we directly use their parameters. 
For pose, we adopt the VPoser~\cite{smplify-x} to encode the 3D body into latent representations $\{f^{3D}_b, f^{3D}_f, f^{3D}_h\}$ for body, face and hands corresponding to $\{\theta^{3D}_b, \theta^{3D}_f, \theta^{3D}_h\}$.
VPoser is a variational auto-encoder trained with large-scale MoCap datasets~\cite{cmu,human3.6m,poseprior}.
Thus it learns a latent space encoding the manifold of the physically plausible pose, and effectively embeds the 3D body pose. 
We concatenate the latent representations, shape parameters and face expression, feed them to two 1024 sized FC layers, and get the 3D human feature $f^{3D}_H=FC_{3D}(\{\beta, \psi, f^{3D}_b, f^{3D}_f, f^{3D}_h\})$.

\subsection{2D-3D Spatial Alignment}
\label{sec:sp-align}
In view of that 2D spatial features lack robustness and may bring in ambiguities, we propose the 2D spatial alignment. 
3D spatial features are more robust, thus we refer them as anchors in the spatial space which describes the manifold of HOI spatial configuration. 
Given the 2D spatial feature $f^{2D}_{sp}$ of a sample, from the train set we randomly sample a positive 3D spatial feature $f^{3D}_{sp+}$ with the same HOI label and a negative feature $f^{3D}_{sp-}$ with non-overlapping HOIs (a person may perform multiple actions at the same time).
For a human-object pair, we use triplet loss~\cite{triplet} to align its 2D spatial feature, \ie, 
\begin{eqnarray}
    \mathcal{L}_{tri} = [d(f^{2D}_{sp}, f^{3D}_{sp+})-d(f^{2D}_{sp}, f^{3D}_{sp-}) + \alpha]_{+}
\label{eq:triplet-loss}
\end{eqnarray}
where $d(\cdot)$ indicates the Euclidean distance, and $\alpha=0.5$ is the margin value.
For 2D samples with the same HOIs but different 2D spatial configurations, this spatial alignment will gather them together in the spatial space. 

\subsection{Joint Body Part Attention Estimation}
\label{sec:att-align}
\begin{figure}[!ht]
	\begin{center}
		\includegraphics[width=0.45\textwidth]{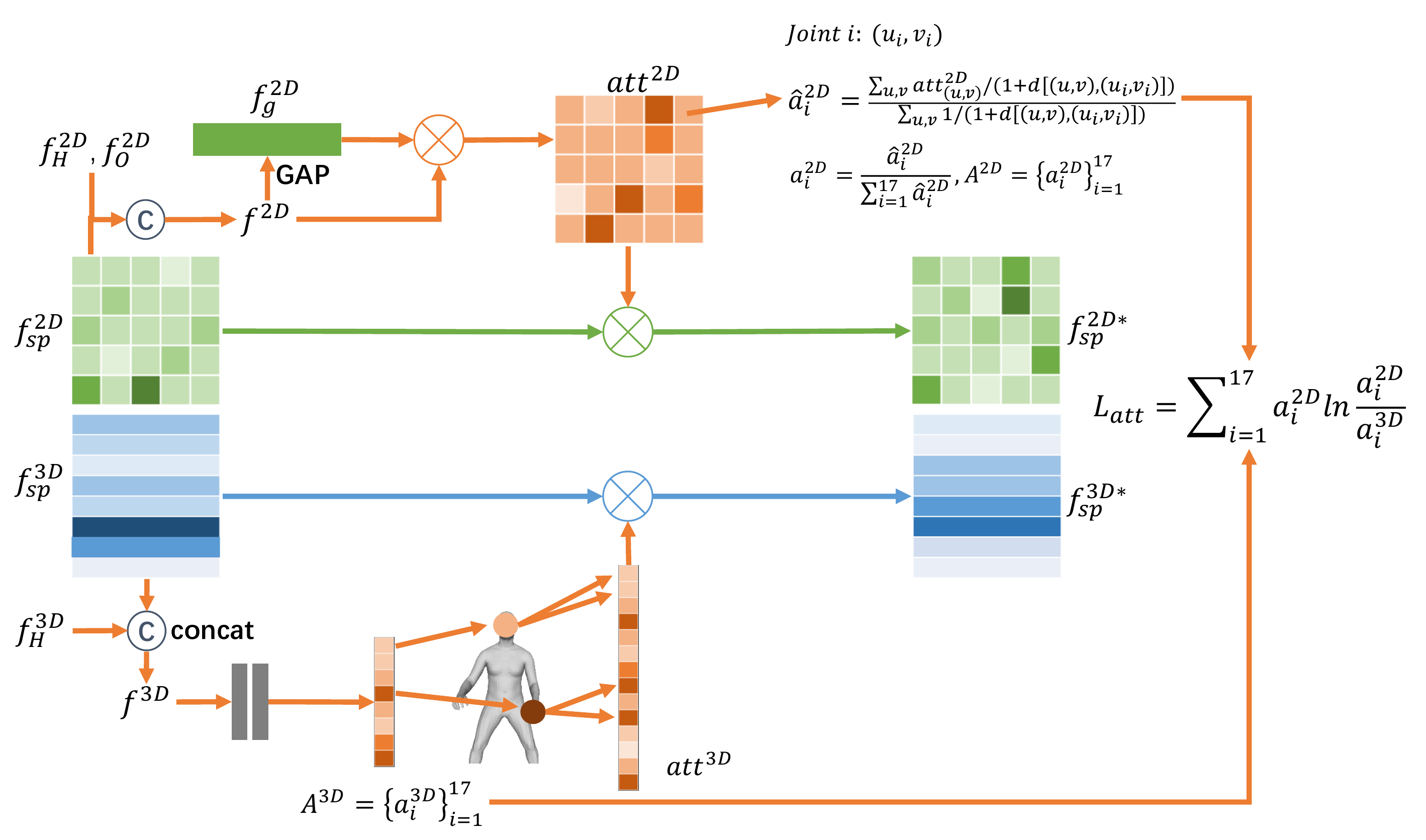}
	\end{center}
	\caption{Body part attention alignment. 
	For 2D, we apply self-attention on $f^{2D}_{sp}$ to generate attention map $att^{2D}$ and 2D part attention $A^{2D}$. 
	For 3D, we use $f^{3D}_{sp}$ to generate 3D part attention $A^{3D}$, and get attention map $att^{3D}$ using the correspondency between point cloud and joints. 
	Finally, we construct consistency loss $\mathcal{L}_{att}$ with $A^{3D}$ and $A^{2D}$. 
	$att^{2D}$ and $att^{3D}$ are used to re-weight and generate $f^{2D*}_{sp}$ and $f^{3D*}_{sp}$.}
	\label{Figure:att-align}
\vspace{-0.5cm}
\end{figure}

Body parts are important in HOI understanding, but not all parts make great contributions in inference. 
Thus, adopting attention mechanism is apparently a good choice.
Different from previous methods~\cite{du2017rpan,Fang2018Pairwise}, we generate body part attention by considering both 2D and 3D clues.
Specifically, we use a part attention consistency loss to conduct self-attention learning, as shown in Fig.~\ref{Figure:att-align}.
With the 2D and 3D features, we can generate two sets of body attention.

\noindent{\bf 2D Attention.} 
We concatenate the input $f^{2D}_H, f^{2D}_O, f^{2D}_{sp}$ to get $f^{2D}$, and apply global average pooling (GAP) to get the global feature vector $f^{2D}_{g}$. 
Then we calculate the inner product $\left<f^{2D}_g, f^{2D}\right>$ and generate the attention map $att^{2D}$ by $att^{2D}=Softmax(\left<f^{2D}_g, f^{2D}\right>)$.
Because 2D pose joints can indicate the part locations, we use joint attention to represent 2D part attention.
If a joint location has high attention, its \textit{neighboring} points should have high attention too.
Thus we can calculate the pose joint attention by summarizing the attentions of its neighboring points.
We represent the attention of 17 pose joints as $A^{2D}=\{a^{2D}_i\}_{i=1}^{17}$, 
\begin{eqnarray}
    \hat{a}^{2D}_{i}=\frac{\sum_{u,v}att^{2D}_{(u, v)}/(1+d[(u, v),(u_i, v_i)])}{\sum_{u,v}1/(1+d[(u, v),(u_i, v_i)])},
\label{eq:2D-att}
\end{eqnarray}
and $a^{2D}_{i}=\frac{\hat{a}^{2D}_i}{\sum_{i=1}^{17}\hat{a}^{2D}_i}$, where $(u, v)$ denotes arbitrary point on attention map $att^{2D}$, $(u_i, v_i)$ indicates the coordinate of the $i$-th joint (calculated by scaling the joint coordinate on image). 
$d[\cdot]$ denotes the Euclidean distance between two points.
Eq.~\ref{eq:2D-att} means: if point $(u, v)$ is far from $(u_i, v_i)$, the attention value of $(u, v)$ contributes less to the attention value of $(u_i, v_i)$; if $(u, v)$ is close to $(u_i, v_i)$, it contributes more.
After the summarizing and normalizing, we finally obtain the attention of $(u_i, v_i)$, \ie $a^{2D}_{i}$. 

\noindent{\bf 3D Attention.}
We use 3D joint attention to represent the 3D body part attention.
Input $f_{sp}^{3D}$ is $[1228\times 384]$ and $f_H^{3D}$ is $[1024]$. 
We first {\bf tile} $f_H^{3D}$ 1228 times to get shape $[1228\times1024]$, then concatenate it with $f_{sp}^{3D}$ to get $f^{3D}$ ($[1228\times 1408]$). 
Then we apply GAP to $f^{3D}$ to get a $[1408]$ tensor, and feed it to two 512 sized FC layers and Softmax, finally obtain the attention for 17 joints, \ie, $A^{3D}=\{a^{3D}_j\}^{17}_{j=1}$.

\noindent{\bf Attention Consistency.}
Whereafter, we operate the attention alignment via an attention consistency loss:
\begin{eqnarray}
    \mathcal{L}_{att} = \sum_{i}^{17} a_{i}^{2D}ln\frac{a_{i}^{2D}}{a_{i}^{3D}}.
\label{eq:att-consistency-loss}
\end{eqnarray}
where $a_{i}^{2D}$ and $a_{i}^{3D}$ are 2D and 3D attentions of the $i$-th joint. $L_{att}$ is the Kullback–Leibler divergence between $A^{2D}$ and $A^{3D}$, which enforces two attention estimators to generate similar \textbf{part importance} and keep the consistency.

Next, in 2D-RN, we multiply $f^{2D}_{sp}$ by $att^{2D}$, \ie the Hadamard product $f^{2D*}_{sp}$=$f^{2D}_{sp} \circ att^{2D}$.
In 3D-RN, we first assign attention to each 3D point in the spatial configuration volume ($n$ points in total).
For human 3D points, we divide them into different sets according to 17 joints, and each set is corresponding to a body part. 
Within the $i$-th set, we \textit{tile} the body part attention $a_{i}^{3D}$ to each point.
For object 3D points, we set all their attention as one.
Because each element of $f^{3D}_{sp}$ is corresponding to a 3D point in the spatial configuration volume, we organize the attentions of both human and object 3D points as $att^{3D}$ of size $n \times 1$, where $n$ is the number of elements in $f^{3D}_{sp}$ (Fig.~\ref{Figure:att-align}).
Thus we can calculate the Hadamard product $f^{3D*}_{sp}$=$f^{3D}_{sp} \circ att^{3D}$.
After the part feature re-weighting, our model can learn to neglect the parts unimportant to the HOI inference.

\subsection{2D-3D Semantic Consistency}
\label{sec:semantic-align}
After the feature extraction and re-weighting, we perform the HOI classification.
All classifiers in each stream are composed of two 1024 sized FC layers and Sigmoids. 
The HOI score of the 2D-RN is $\mathcal{S}^{2D}=(s^{2D}_H + s^{2D}_O) \circ s^{2D}_{sp}$, where $s^{2D}_H, s^{2D}_O, s^{2D}_{sp}$ are the scores of human, object and spatial stream.
$\mathcal{S}^{3D}=s^{3D}_H + s^{3D}_{sp}$ indicates the final prediction of the 3D-RN. 
To maintain the semantic consistency of 2D and 3D representations, \ie they should make the same prediction for the same sample, we construct:
\begin{eqnarray}
    \mathcal{L}_{sem} = \sum_{i}^{m} ||\mathcal{S}_{i}^{2D}-\mathcal{S}_{i}^{3D}||_2,
\label{eq:2d-3d-ouput-consistency-loss}
\end{eqnarray}
where $m$ is the number of HOIs.

\noindent{\bf Multiple HOI Inferences.}
Moreover, we concatenate the features from the last FC layers in 2D-RN and 3D-RN as $f^{joint}$ (early fusion), and make the third classification to obtain the score $\mathcal{S}^{joint}$. The joint classifier is also composed of two 1024 sized FC layers and Sigmoids.
The multi-label classification cross-entropy losses are expressed as $\mathcal{L}^{2D}_{cls}, \mathcal{L}^{3D}_{cls}, \mathcal{L}^{joint}_{cls}$.
Thus, the total loss of DJ-RN is:
\begin{eqnarray}
\begin{split}
        \mathcal{L}_{total} = \lambda_{1}\mathcal{L}_{tri} + \lambda_{2}\mathcal{L}_{att} + \lambda_{3}\mathcal{L}_{sem} + \lambda_{4}\mathcal{L}_{cls},
\end{split}
\label{eq:total-loss}
\end{eqnarray}
where $\mathcal{L}_{cls}$=$\mathcal{L}^{2D}_{cls}$+$\mathcal{L}^{3D}_{cls}$+$\mathcal{L}^{joint}_{cls}$, and we set $\lambda_1$=0.001, $\lambda_2$ =0.01, $\lambda_3$=0.01, $\lambda_4$=1 in experiments.
The final score is
\begin{eqnarray}
\label{eq:final-score}
    \mathcal{S} = \mathcal{S}^{2D} + \mathcal{S}^{3D} + \mathcal{S}^{joint}.
\end{eqnarray}

\section{Experiment}
In this section, we first introduce the adopted datasets and metrics. Then we describe the detailed implementation of DJ-RN. Next, we compare DJ-RN with the state-of-the-art on HICO-DET~\cite{hicodet} and Ambiguous-HOI. 
At last, ablation studies are operated to evaluate modules in DJ-RN.

\subsection{Ambiguous-HOI}
\label{sec:new-benchmark}
\begin{figure}
    \centering
    \includegraphics[width=0.45\textwidth]{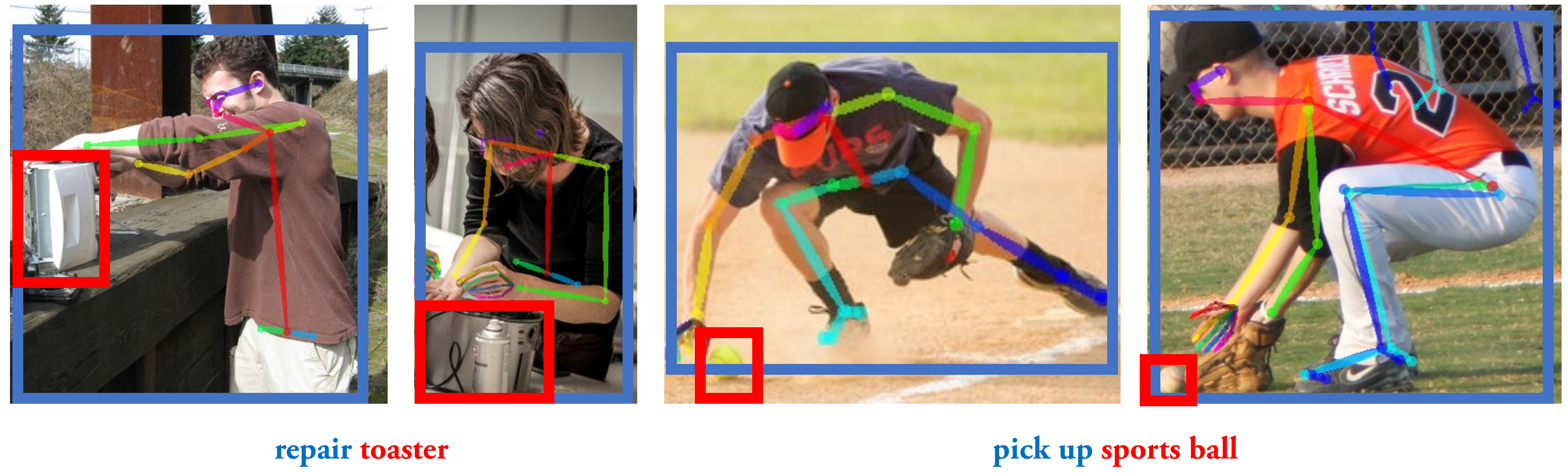}
    \caption{Ambiguous samples from Ambiguous-HOI.}
    \label{fig:samples}
    \vspace{-0.5cm}
\end{figure}
Existing benchmarks mainly focus on evaluating generic HOIs, but not to specially examine the ability to process 2D pose and appearance ambiguities. Hence, we propose a new benchmark named Ambiguous-HOI.
Ambiguous-HOI consists of hard examples collected from the test set of HICO-DET~\cite{hicodet}, and other whole datasets such as V-COCO~\cite{vcoco}, OpenImage~\cite{openimages}, HCVRD~\cite{hcvrd} and Internet images.
We choose HOI categories from HICO-DET~\cite{hicodet} for its well-designed verbs and objects.
For Internet images, we labeled the HOIs according to HICO-DET setting.
The 2D pose and spatial configuration ambiguities are mainly considered in the selection.
First, we put all images and corresponding labels in a candidate pool and manually choose some template 2D pose samples for each HOI. 
Then we use Procrustes transformation~\cite{borg1997modern} to align the 2D pose of samples to the templates.
Next, we cluster all samples to find the samples far from the cluster center and repeat clustering according to different templates.
The mean distance between a sample and multiple cluster centers is recorded as reference.
Meanwhile, we train an MLP taking the 2D pose and spatial map as inputs on HICO-DET train set. 
Then we use it as an \textit{ambiguity probe} to find the most easily misclassified samples.
Combining the above two references, we finally select 8,996 images with 25,188 annotated human-object pairs. 
Ambiguous-HOI finally includes 87 HOI categories, consisting of 48 verbs and 40 object categories from HICO-DET~\cite{hicodet}. Some sample are shown in Fig.~\ref{fig:samples}.

\subsection{Dataset and Metric}
\noindent{\bf Dataset.} We adopt the widely-used HOI benchmark HICO-DET~\cite{hicodet} and our novel Ambiguous-HOI. HICO-DET~\cite{hicodet} is an instance-level benchmark consisting of 47,776 images (38,118 for training and 9,658 for testing) and 600 HOI categories. It contains 80 object categories from COCO~\cite{coco}, 117 verbs and more than 150k annotated HOI pairs. 

\noindent{\bf Metric.} We use mAP metric from \cite{hicodet} for two benchmarks: true positive need to contain accurate human and object locations (box IoU with reference to the ground truth box is larger than 0.5) and accurate interaction/verb classification. 

\subsection{Implementation Details}
For 3D body recovery, we first use OpenPose~\cite{openpose} to detect the 2D pose of body, face and hands. Then we feed them with the image to SMPLify-X~\cite{smplify-x} to get 3D body. 
Since cases with severe occlusion might fail the 3D recovery, we only recover 3D bodies for those which at least includes detected 2D head, pelvis, one shoulder and one hip joints.
For the rest, we assign them the body with standard template 3D pose, \ie generated by setting all SMPL-X parameters to zero.
Sometimes the recovered body can be implausible, \ie ``monsters''. 
To exclude them, we use VPoser\cite{smplify-x} to extract the latent embedding of every recovered 3D body.
With the \textit{mean latent embedding} of the generated 3D body from HICO-DET train set as a reference, we assume that the farthest 10\% embeddings from the mean embedding are ``monsters''.
At last, 81.2\% of the annotated instances are assigned with SMPLify-X~\cite{smplify-x} generated mesh, and we assign standard templates for the rest to avoid importing noise.

For feature extraction, we use COCO~\cite{coco} pre-trained ResNet-50~\cite{resnet} in 2D-RN. 
In 3D-RN, we first train a PointNet~\cite{pointnet} to segment the human and object points in 3D volume, and then use it to extract the 3D local feature of volume. 
The PointNet is trained for 10K iterations, using SGD with learning rate of 0.01, momentum of 0.9 and batch size of 32. 
In spatial alignment, we adopt the triplet loss with semi-hard sampling, \ie, for a sample, we only calculate the loss for its nearest negative and farthest positive samples in the same mini-batch with respect to their Euclidean distances. 
In joint training, we train the whole model for 400K iterations, using SGD with momentum of 0.9, following cosine learning rate restart~\cite{cosinelr} with initial learning rate of 1e-3.
For a fair comparison, we use object detection from iCAN~\cite{gao2018ican}.
We also adopt the Non-Interaction Suppression (NIS) and Low-grade Instance Suppression (LIS)~\cite{interactiveness} in inference. 
The interactiveness model from ~\cite{interactiveness} is trained on HICO-DET train set only. 
The thresholds of NIS are 0.9 and 0.1 and LIS parameters follow \cite{interactiveness}.

\begin{table}
\centering
\resizebox{0.48\textwidth}{!}{
\begin{tabular}{l  c  c  c  c  c  c  }
\hline
         & \multicolumn{3}{c}{Default}  &\multicolumn{3}{c}{Known Object} \\
Method         & Full & Rare & Non-Rare  & Full & Rare & Non-Rare \\
\hline
\hline
Shen~\etal~\cite{Shen2018Scaling}        & 6.46  & 4.24  & 7.12  & - & - & -\\
HO-RCNN~\cite{hicodet}                   & 7.81  & 5.37  & 8.54  & 10.41 & 8.94  & 10.85\\
InteractNet~\cite{Gkioxari2017Detecting} & 9.94  & 7.16  & 10.77 & - & - & -\\
GPNN~\cite{qi2018learning}               & 13.11 & 9.34  & 14.23 & - & - & -\\
iCAN~\cite{gao2018ican}                  & 14.84 & 10.45 & 16.15 & 16.26 & 11.33 & 17.73\\
Interactiveness~\cite{interactiveness}   & 17.03 & 13.42 & 18.11 & 19.17 & 15.51 & 20.26\\
No-Frills~\cite{NoFrills}                & 17.18 & 12.17 & 18.68 & - & - & -\\
PMFNet~\cite{shangke}                    & 17.46 & 15.65 & 18.00 & 20.34 & 17.47 & 21.20 \\
Julia~\etal~\cite{peyre2018detecting}    & 19.40 & 14.60 & 20.90 & - & - & -\\
\hline
$\mathcal{S}^{2D}$             & 19.98 & 16.97 & 20.88 & 22.56 & 19.48 & 23.48 \\ 
$\mathcal{S}^{3D}$             & 12.41 & 13.08 & 12.21 & 16.95 & 17.74 & 16.72 \\ 
$\mathcal{S}^{Joint}$          & 20.61 & 17.01 & 21.69 & 23.21 & 19.66 & 24.28 \\ 
\hline           
DJ-RN             & \textbf{21.34} & \textbf{18.53} & \textbf{22.18} & \textbf{23.69} & \textbf{20.64} & \textbf{24.60}\\ 
\hline
\end{tabular}}
\caption{Results comparison on HICO-DET~\cite{hicodet}. }
\label{tab:hico-det}
\vspace{-0.1cm}
\end{table}

\subsection{Results and Comparisons}
\noindent{\bf HICO-DET.}
We demonstrate our quantitative results in Tab.~\ref{tab:hico-det}, compared with state-of-the-art methods~\cite{Shen2018Scaling,hicodet,Gkioxari2017Detecting,qi2018learning,gao2018ican,interactiveness,NoFrills,shangke,peyre2018detecting}. The evaluation follows the settings in HICO-DET\cite{hicodet}: Full(600 HOIs), Rare(138 HOIs) and Non-Rare(462 HOIs) in Default and Known Object mode. 
We also evaluate different streams in our model, \ie 2D ($\mathcal{S}^{2D}$), 3D ($\mathcal{S}^{3D}$) and Joint ($\mathcal{S}^{joint}$).
Our 2D-RN has a similar multi-stream structure, object detection and backbone following HO-RCNN~\cite{hicodet}, iCAN~\cite{gao2018ican}, Interactiveness~\cite{interactiveness} and PMFNet~\cite{shangke}.
With {\bf joint learning}, 2D-RN ($\mathcal{S}^{2D}$) directly outperforms above methods with {\bf 13.53, 6.50, 4.31, 3.88} mAP on Default Full set. 
This strongly proves the effectiveness of the consistency tasks in joint learning.
Meanwhile, 3D-RN ($\mathcal{S}^{3D}$) achieves 12.41 mAP on Default Full set and shows obvious complementarity for 2D-RN. 
Especially, 3D performs better on Rare set than Non-Rare set.
This suggests that 3D representation has much \textit{weaker data-dependence} than 2D representation and is less affected by the long-tail data distribution.
Joint learning ($\mathcal{S}^{Joint}$) performs better than both 2D and 3D, achieving 20.61 mAP, while unified DJ-RN (late fusion) finally achieves 21.34 mAP, which outperforms the latest state-of-the-art~\cite{peyre2018detecting} with 1.94 mAP. 
Facilitated by the detailed 3D body information, we achieve \textbf{21.71} mAP on 356 \textit{hand-related} HOIs, which is higher than the 21.34 mAP on 600 HOIs.  

\begin{figure}[!ht]
	\begin{center}
		\includegraphics[width=0.45\textwidth]{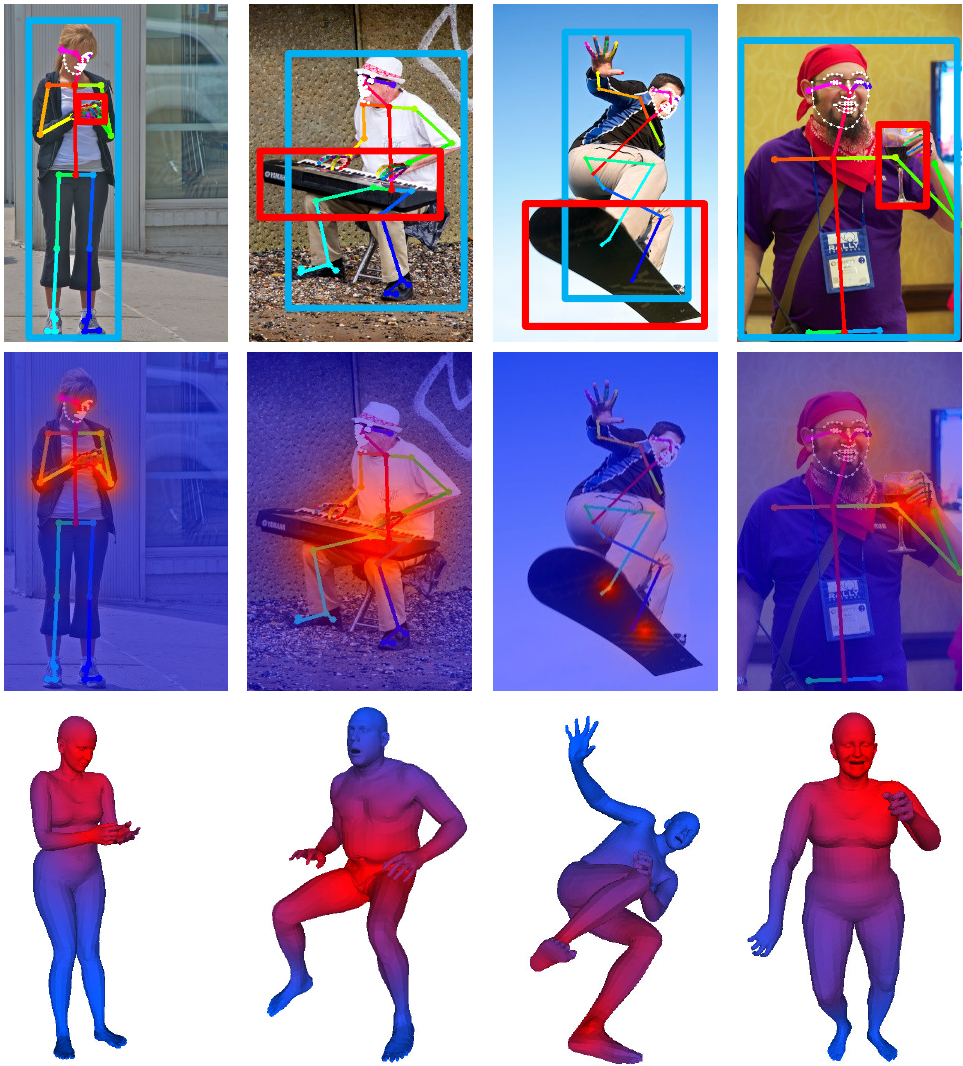}
	\end{center}
	\caption{Visualized attention. Three rows are images, 2D and 3D attentions respectively. Red indicates high attention and blue is the opposite. 2D attention is in line with 3D attention, and they both capture reasonable part attentions for various HOIs.}
	\label{Figure:part-attention}
\vspace{-0.1cm}
\end{figure}

\begin{table}
\centering
\resizebox{0.3\textwidth}{!}{
\begin{tabular}{l c c}
\hline
Method         & Ambiguious-HOI \\
\hline
\hline
iCAN~\cite{gao2018ican}                   & 8.14\\
Interactiveness~\cite{interactiveness}    & 8.22\\
Julia~\etal~\cite{peyre2018detecting}     & 9.72 \\
\hline
DJ-RN                                     & \textbf{10.37} \\
\hline
\end{tabular}}
\caption{Results comparison on Ambiguous-HOI.} 
\label{tab:confusing}
\vspace{-0.1cm}
\end{table}

\begin{table}
\centering
\resizebox{0.48\textwidth}{!}{
\begin{tabular}{l  c  c  c  c  c  c}
\hline
         & \multicolumn{3}{c}{Default}  &\multicolumn{3}{c}{Known Object} \\
Method  & Full & Rare & Non-Rare  & Full & Rare & Non-Rare \\
\hline
DJ-RN             & \textbf{21.34} & \textbf{18.53} & \textbf{22.18} & \textbf{23.69} & \textbf{20.64} & \textbf{24.60}\\ 

\hline
3D Pose                 & 20.42 & 16.88 & 21.47 & 22.95 & 19.48 & 23.99 \\ 
Point Cloud             & 20.05 & 16.52 & 21.10 & 22.61 & 19.11 & 23.66 \\ 
\hline
w/o Face                & 21.02 & 17.56 & 22.05 & 23.48 & 19.80 & 24.58 \\ 
w/o Hands               & 20.83 & 17.36 & 21.87 & 23.40 & 19.99 & 24.41 \\ 
w/o Face \& Hands       & 20.74 & 17.36 & 21.75 & 23.33 & 19.82 & 24.37 \\ 
\hline
w/o Volume Block        & 20.34 & 17.19 & 21.28 & 22.97 & 19.94 & 23.87 \\
w/o 3D Body Block       & 20.01 & 16.14 & 21.17 & 22.73 & 18.88 & 23.88 \\
\hline
w/o $\mathcal{L}_{att}$ & 20.70 & 16.56 & 21.93 & 23.32 & 19.13 & 24.57 \\
w/o $\mathcal{L}_{tri}$ & 20.83 & 17.66 & 21.77 & 23.50 & 20.31 & 24.45 \\
w/o $\mathcal{L}_{sem}$ & 20.80 & 17.51 & 21.78 & 23.45 & 20.27 & 24.39 \\
\hline
\end{tabular}}
\caption{Results of ablation studies.}
\label{tab:ablation}
\vspace{-0.3cm}
\end{table}

\noindent{\bf Ambiguous-HOI.}
\label{sec:confusing-hoi-test}
To further evaluate our method, we conduct an experiment on the proposed Ambiguous-HOI.
We choose methods~\cite{gao2018ican,interactiveness,peyre2018detecting} with open-sourced code as baselines.
All models are trained on HICO-DET train set and achieve respective best performances. 
To test the ability of disambiguation and generalization, we directly test all models on Ambiguous-HOI. 
Ambiguous-HOI is much more difficult, thus all methods get relatively low scores (Tab.~\ref{tab:confusing}). 
DJ-RN outperforms previous method by 0.65, 2.15 and 2.23 mAP.
This strongly verifies the advantage of our joint representation.

\noindent{\bf Visualizations.}
\begin{figure}[!ht]
	\begin{center}
		\includegraphics[width=0.45\textwidth]{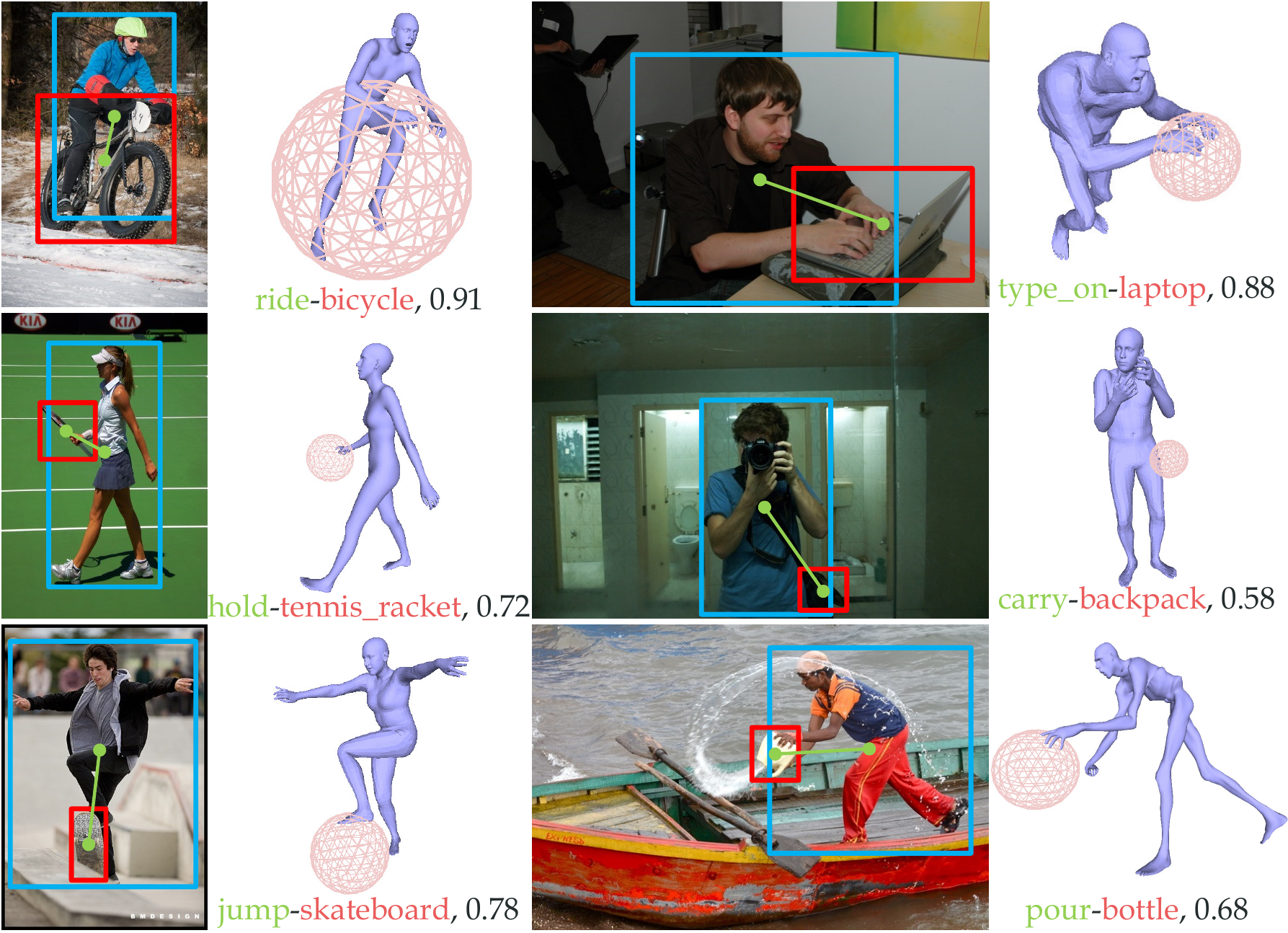}
	\end{center}
	\caption{Visualized results and the corresponding 3D volumes.}
	\label{Figure:visualization-hoi}
\vspace{-0.5cm}
\end{figure}
We visualize the part attention in Fig.~\ref{Figure:part-attention}. We can find that two kinds of attention are aligned well and both capture essential parts for various HOIs. 
We also visualize HOI predictions paired with estimated 3D spatial configuration volumes in Fig.~\ref{Figure:visualization-hoi}. Our method performs robustly in HOI inference and 3D spatial configuration estimation.

\noindent{\bf Time Complexity.}
2D-RN has similar complexity with iCAN~\cite{gao2018ican} and Interactiveness~\cite{interactiveness}. 
3D-RN is very efficient because of the {\bf pre-extracted features} (about 50 FPS).
SMPLif-X runs with GPU acceleration is about 5 FPS.

\subsection{Ablation Study}
We evaluate different components of our method on HICO-DET. The results are shown in Tab.~\ref{tab:ablation}.

\noindent{\bf 3D Formats.} 
Using 3D pose or point cloud for 3D body block in 3D-RN performs worse than VPoser embedding.

\noindent{\bf 3D Human Inputs.}
Without detailed face and hand shape, DJ-RN shows obvious degradation, especially DJ-RN without hand shape.
Because about 70\% verbs in HICO-DET are hand-related, which is consistent with daily experience.

\noindent{\bf Blocks.} Without volume or body block in 3D-RN hurts the performance with 1.00 and 1.33 mAP.

\noindent{\bf Losses:}
Without $\mathcal{L}_{att}$, $\mathcal{L}_{tri}$ and $\mathcal{L}_{sem}$, the performance degrades 0.64, 0.51 and 0.54 mAP.

\section{Conclusion}
In this paper, we propose a novel 2D-3D joint HOI representation learning paradigm, DJ-RN. 
We first represent the HOI in 3D with detailed 3D body and estimated object location and size.
Second, a 2D Representation Network and a 3D Representation Network are proposed to extract multi-modal features. 
Several cross-modal consistency tasks are finally adopted to drive the joint learning.
On HICO-DET and our novel benchmark Ambiguous-HOI, DJ-RN achieves state-of-the-art results.

{\small
\paragraph{Acknowledgement:} This work is supported in part by the National Key R\&D Program of China, No. 2017YFA0700800, National Natural Science Foundation of China under Grants 61772332 and Shanghai Qi Zhi Institute. 
}

{\small
\bibliographystyle{ieee_fullname}
\bibliography{egbib}
}

\clearpage

\onecolumn
\begin{appendices}

\begin{figure}[!ht]
	\begin{center}
		\includegraphics[width=0.9\textwidth]{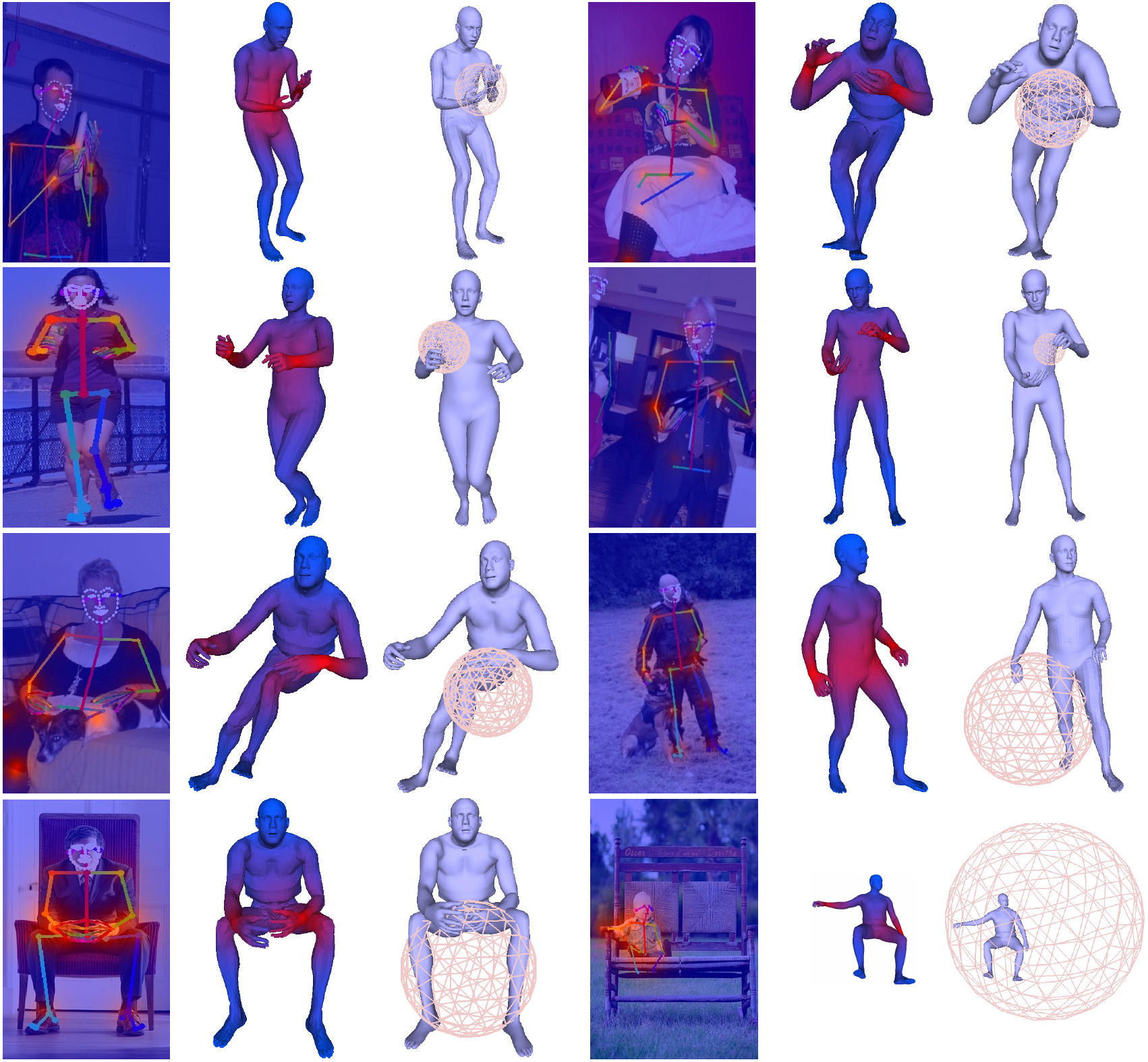}
	\end{center}
	\caption{Visualized human body part attentions and estimated 3D spatial configuration volumes.}
	\label{Figure:vis1}
\end{figure}

\section{Visualized Results}
We report more 2D and 3D visualized body part attentions and estimated 3D spatial configuration volumes in Fig.\ref{Figure:vis1}.
From the results we can find that, our method can well handle various human-object interactions, \eg, from the small bottle to large chair, from local action ``hold something'' to whole body action ``sit on something''.
Meanwhile, our method can also well estimated the interacted object size and location, \eg from the simple banana to the hard chair. An interesting case is in the fourth row, the image includes a baby who sits in a big chair. We can find that our method successfully estimates appropriate chair size relative to the baby, and the location is also accurate which completely covers the baby in 3D volume.

\begin{table}[!ht]
    \centering
    \begin{tabular}{c|c|c|c|c|c|c|c}
        \hline
        Object category & Ratio & $\gamma^{min}$ & $\gamma^{max}$ &Object category & Ratio & $\gamma^{min}$ & $\gamma^{max}$\\
        \hline
        airplane        & 195.640 &  1.0 & 1.0 & apple           & 0.205  & 1.0  & 1.0  \\
        backpack        & 0.769   &  1.0 & 1.0 & banana          & 0.385  & 1.0  & 1.0  \\
        baseball\_bat   & 2.564   &  1.0 & 1.0 & baseball\_glove & 0.769  & 1.0  & 1.0  \\
		bear            & 5.128	  &  0.7 & 1.3 & bed             & 5.128  & 1.0  & 1.0  \\
        bench           & 3.128   &  1.0 & 1.0 & bicycle         & 1.051  & 1.0  & 1.0  \\
        bird            & 0.718   &  0.7 & 1.3 & boat            & 12.821 & 1.0  & 1.0  \\
        book			& 0.590   &  1.0 & 1.0 & bottle          & 0.769  & 1.0  & 1.0  \\
        bowl            & 0.487   &  1.0 & 1.0 & broccoli        & 0.256  & 1.0  & 1.0  \\
        bus 			& 3.590	  &  1.0 & 1.0 & cake		     & 0.462  & 0.8  & 1.2  \\
        car				& 9.744	  &  1.0 & 1.0 & carrot		     & 0.103  & 1.0  & 1.0  \\
        cat				& 1.179	  &  1.0 & 1.0 & cell\_phone     & 0.333  & 1.0  & 1.0  \\
        chair     		& 1.103	  &  1.0 & 1.0 & clock 		     & 0.718  & 0.7  & 1.3  \\
        couch 			& 4.744	  &  1.0 & 1.0 & cow 		     & 4.359  & 0.8  & 1.2  \\
        cup				& 0.564   &  1.0 & 1.0 & dining\_table   & 4.615  & 1.0  & 1.0  \\
        dog 			& 1.231	  &  1.0 & 1.0 & donut		     & 0.128  & 1.0  & 1.0  \\
        elephant	    & 7.436	  &  0.8 & 1.2 & fire\_hydrant   & 0.308  & 1.0  & 1.0  \\
        fork 			& 0.410	  &  1.0 & 1.0 & frisbee		 & 0.513  & 0.6  & 1.4  \\
        giraffe			& 10.769  &  0.8 & 1.2 & hair\_drier     & 0.513  & 1.0  & 1.0  \\
        handbag         & 1.385	  &  0.8 & 1.2 & horse           & 5.385  & 0.8  & 1.2  \\
        hot\_dog        & 0.385	  &  1.0 & 1.0 & keyboard        & 0.641  & 1.0  & 1.0  \\
        kite   			& 2.051	  &  0.5 & 1.5 & knife			 & 0.410  & 1.0  & 1.0  \\
        laptop			& 0.846	  &  1.0 & 1.0 & microwave       & 1.154  & 0.8  & 1.2  \\
        motorcycle      & 3.949	  &  1.0 & 1.0 & mouse 		     & 0.256  & 1.0  & 1.0  \\
        orange			& 0.179	  &  1.0 & 1.0 & oven 			 & 1.538  & 0.8  & 1.2  \\
        parking\_meter  & 4.103	  &  0.8 & 1.2 & person 		 & 4.487  & 0.8  & 1.2  \\
        pizza			& 0.769	  &  1.0 & 1.0 & potted\_plant   & 0.590  & 0.8  & 1.2  \\
        refrigerator    & 4.231	  &  1.0 & 1.0 & remote  		 & 0.513  & 1.0  & 1.0  \\
        sandwich	    & 0.359	  &  1.0 & 1.0 & scissors		 & 0.385  & 0.8  & 1.2  \\
        sheep 			& 3.333	  &  0.8 & 1.2 & sink			 & 1.282  & 1.0  & 1.0  \\
        skateboard      & 1.821	  &  1.0 & 1.0 & skis			 & 3.846  & 1.0  & 1.0  \\
        snowboard       & 3.949	  &  1.0 & 1.0 & spoon		     & 0.410  & 1.0  & 1.0  \\
        sports\_ball    & 1.795	  &  0.8 & 1.2 & stop\_sign      & 3.923  & 0.6  & 1.4  \\
        suitcase        & 1.615	  &  1.0 & 1.0 & surfboard       & 6.231  & 1.0  & 1.0  \\
        teddy\_bear     & 2.462	  &  1.0 & 1.0 & tennis\_racket  & 1.897  & 1.0  & 1.0  \\
        tie				& 1.308	  &  1.0 & 1.0 & toaster 		 & 0.641  & 0.8  & 1.2  \\
        toilet			& 1.103	  &  1.0 & 1.0 & toothbrush      & 0.436  & 1.0  & 1.0  \\
        traffic\_light  & 0.651	  &  0.6 & 1.4 & train 		     & 512.82 & 1.0  & 1.0  \\
        truck			& 5.385	  &  1.0 & 1.0 & tv			     & 1.821  & 0.7  & 1.3  \\
        umbrella        & 2.949	  &  1.0 & 1.0 & vase			 & 0.846  & 0.8  & 1.2  \\
        wine\_glass     & 0.462	  &  1.0 & 1.0 & zebra			 & 6.154  & 0.8  & 1.2  \\
        \hline
    \end{tabular}
    \caption{Object prior size ratio relative to the human shoulder width and object prior depth regularization factor $\Gamma=\{\gamma^{min}_i, \gamma^{max}_i\}_{i=1}^{80}$.}
    \label{tab:object-prior}
\end{table}

\section{Prior Object Size and Depth Regularization Factor}
When estimating the 3D spatial configuration volume, we utilize the prior object information collected from volunteers.
The lists of the prior size and depth regularization factors of COCO 80 object~\cite{coco} in the HICO-DET~\cite{hicodet} are shown in Tab.~\ref{tab:object-prior}.

\section{Volunteer Backgrounds}
About 50 volunteers took part in our prior object information collection.
The volunteer backgrounds are detailed in Tab.~\ref{table:volunteer}. 
\begin{table*}[!ht]
 	\begin{center}
 		\begin{tabular}{p{60pt}p{410pt}}
 		\hline
 	    Background & Content \\
 	    \hline
 	    Age   & 18-20 (9, 18\%), 21-25 (23, 46\%), 26-30 (18, 36\%) \\
 	    Education & High School (25, 50\%), Bachelor (13, 26\%), Master (9, 18\%), PhD (3, 6\%)\\
        Major & Law (2, 4\%), Agriculture (9, 18\%), Economics (1, 2\%), Education (9, 18\%), Medicine (5, 10\%), Engineering (24, 48\%)\\
 	    \hline
 		\end{tabular}
 	\end{center}
 	\caption{Volunteer backgrounds.}
 	\label{table:volunteer}
\end{table*}
\begin{table}[!ht]
    \centering
    \begin{tabular}{c|c|c|c|c|c}
        \hline
        Object & Verb & Human-Object & Object & Verb & Human-Object\\
        \hline
        surfboard    &    load       &   46 & carrot       & stir     &   19\\
        orange       &    wash       &   123& wine\_glass  & wash     &   6\\
        bottle       &   open        &   10 &car           & direct   &   7\\
        bus          &   inspect     &   16 &frisbee       & spin     &   202\\
        apple        &   wash        &   48 &bowl          & lick     &   5\\
        spoon        &   wash        &   73 &boat          & wash     &   23\\
        person       &   teach       &   13 &bicycle       & wash     &   128\\
        pizza        &   buy         &   7  &orange        & eat      &   5\\
        toilet       &   stand\_on   &   80 &carrot        & cook     &   57\\
        elephant     &   hop\_on     &   5  &sports\_ball  & pick\_up &   236\\
        cat          &   chase       &   7  &pizza         & slide    &   18\\
        pizza        &   smell       &   15 &oven          & repair   &   87\\
        sports\_ball &   catch       &   270&person        & stab     &   175\\
        bear         &   feed        &   7  &dining\_table & clean    &   7\\
        cow          &   kiss        &   5  &car           & board    &   10\\
        elephant     &   wash        &   42 &dog           & chase    &   121\\
        giraffe      &   ride        &   20 &vase          & paint    &   64\\
        backpack     &   open        &   7  &surfboard     & sit\_on  &   210\\
        cat          &   kiss        &   136&knife         & stick    &   224\\
        dog          &   groom       &   21 &horse         & feed     &   6\\
        giraffe      &   pet         &   8  &hair\_drier   & repair   &   5\\
        dog          &   wash        &   229&umbrella      & open     &   28\\
        horse        &   load        &   209&teddy\_bear   & kiss     &   49\\
        boat         &   exit        &   268&train         & exit     &   63\\
        person       &   hug         &   211&car           & park     &   13\\
        backpack     &   inspect     &   211&sheep         & wash     &   9\\
        sheep        &   pet         &   13 &motorcycle    & wash     &   152\\
        toaster      &   repair      &   13 &bed           & clean    &   6\\
        sports\_ball &   hold        &   221&skis          & wear     &   264\\
        bus          &   wash        &   24 &sports\_ball  & block    &   22\\
        dog          &   feed        &   15 &train         & load     &   100\\
        bird         &   chase       &   61 &airplane      & exit     &   23\\
        book         &   carry       &   67 &dog           & run      &   233\\
        kite         &   assemble    &   23 &baseball\_bat & carry    &   17\\
        fork         &   wash        &   7  &couch         & carry    &   13\\
        bus          &   load        &   31 &fire\_hydrant & open     &   5\\
        person       &   greet       &   221&cow           & ride     &   18\\
        giraffe      &   kiss        &   49 &dog           & straddle &   98\\
        refrigerator &   hold        &   127&car           & inspect  &   29\\
        airplane     &   inspect     &   52 &parking\_meter& pay      &   12\\
        car          &   wash        &   45 &cow           & walk     &   50\\
        bird         &   release     &   78 &horse         & hop\_on  &   6\\
        toilet       &   clean       &   5  &elephant      & hose     &   134\\
        kite         &   inspect     &   238\\
        \hline
    \end{tabular}
    \caption{The selected HOIs of Ambiguous-HOI. ``H-O'' is the number of the annotated human-object pairs of the corresponding HOI.}
    \label{tab:ambiguous_num}
\end{table}
\begin{figure}
	\begin{center}
		\includegraphics[width=0.6\textwidth]{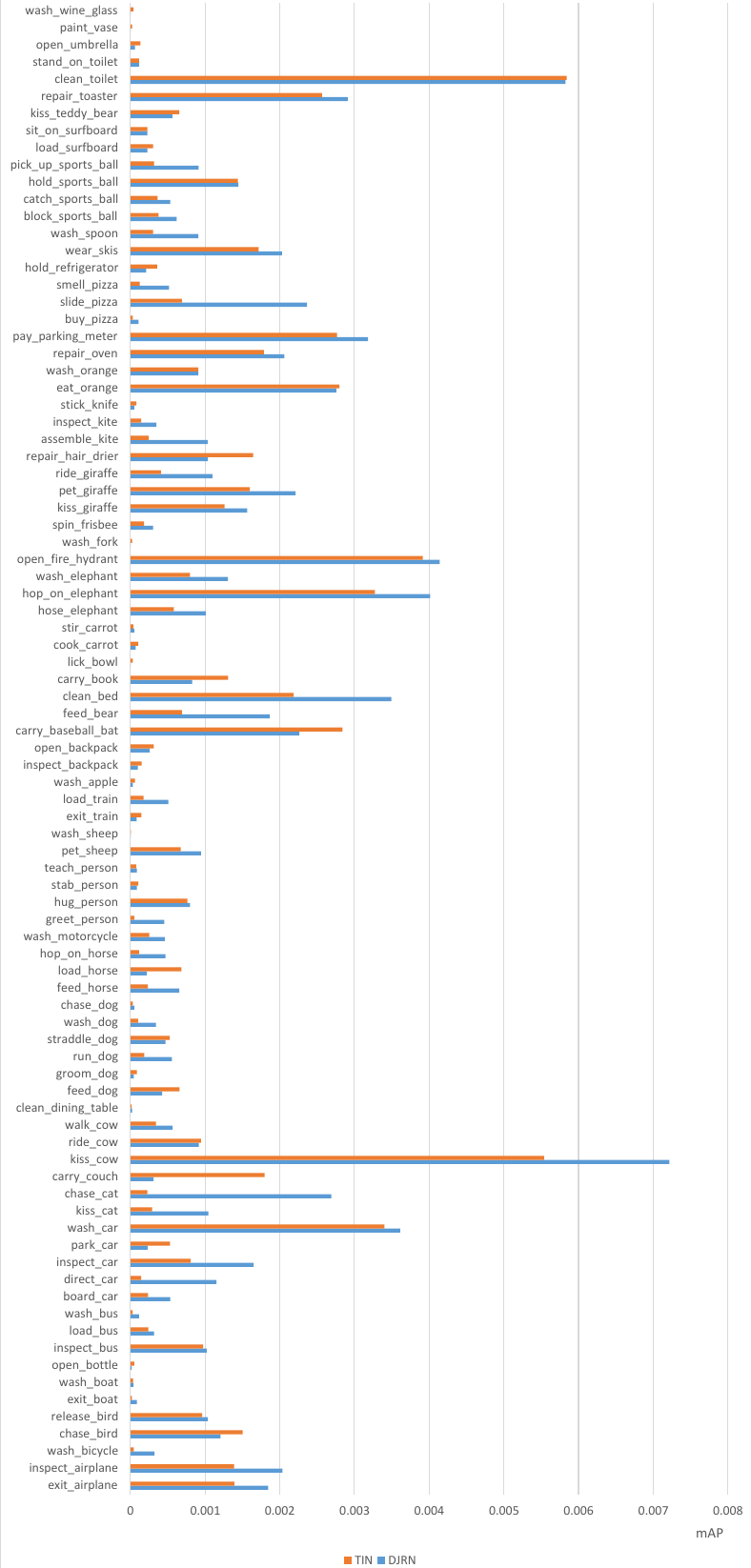}
	\end{center}
    \caption{Performance comparison between our method and TIN~\cite{interactiveness} on Ambiguous-HOI.}
    \label{fig:ambiguous-comp}
\end{figure}

\begin{figure}
	\begin{center}
		\includegraphics[width=0.6\textwidth]{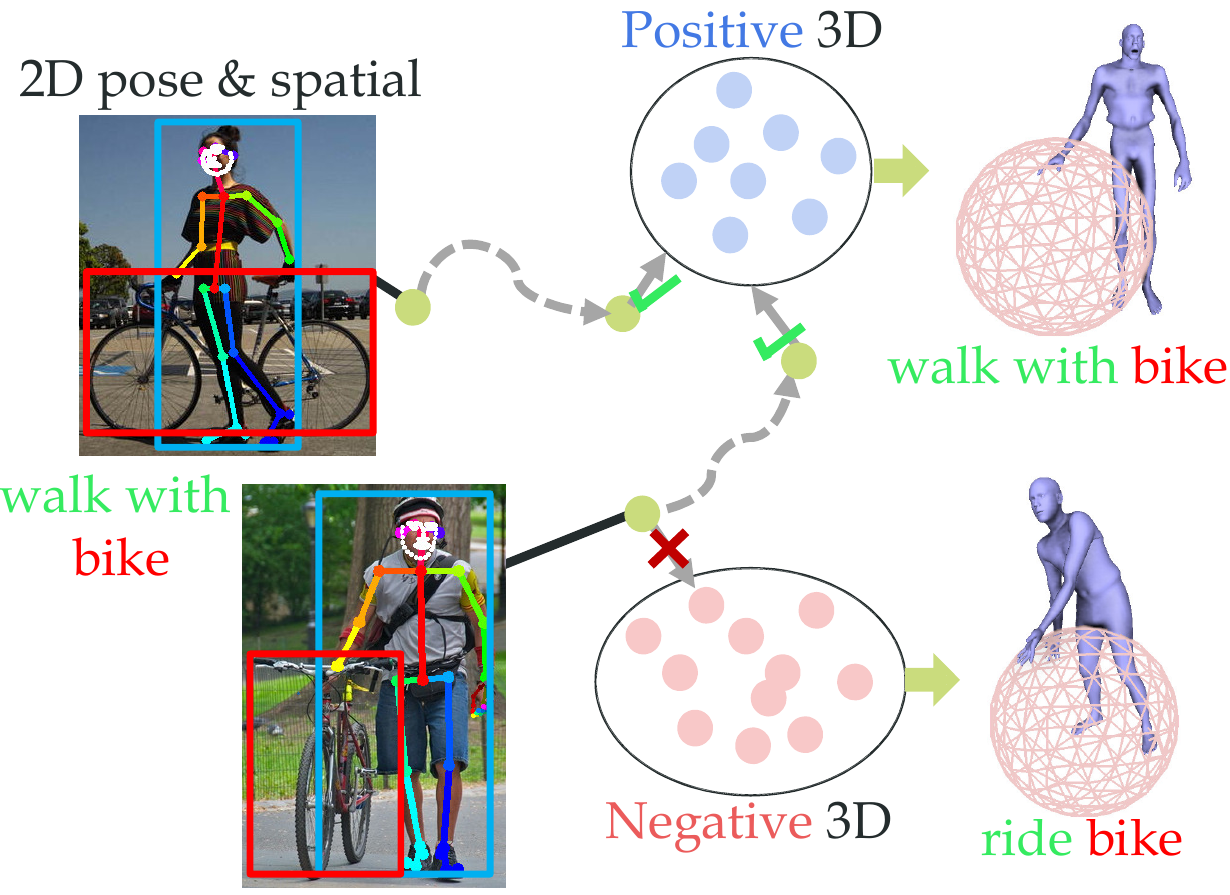}
	\end{center}
    \caption{Illustration of the spatial alignment.}
    \label{fig:sp-align}
\end{figure}

\section{Spatial Alignment Illustration}
In addition, we give a visualized illustration of the spatial alignment between 2D and 3D spatial features in latent space.

\section{Characteristics of Ambiguous-HOI}
Ambiguous-HOI contains 87 kinds of HOIs, which consists of 40 verb categories and 48 object categories.
The detailed statistics of Ambiguous-HOI are shown in Tab.~\ref{tab:ambiguous_num}, which includes the selected object categories, verbs and the number of annotated human-object pairs of each HOI.
We also illustrate the detailed comparison between our method and TIN~\cite{interactiveness} on Ambiguous-HOI in Fig.~\ref{fig:ambiguous-comp}.
We can find that our DJ-RN outperforms TIN on various HOIs and shows the effectiveness of the detailed 2D-3D joint HOI representation. 

\end{appendices}

\end{document}